%% file: root.tex
\documentclass[lettersize,journal]{IEEEtran}  % Comment this line out if you need a4paper
\usepackage{graphicx} % for pdf, bitmapped graphics files
\usepackage{svg}
\usepackage{booktabs}
\usepackage{stfloats}
\usepackage{amsmath, amsfonts} % assumes amsmath package installed
\usepackage{amssymb}  % assumes amsmath package installed
\usepackage{babel}
\usepackage{array}
\usepackage[caption=false,font=normalsize,labelfont=sf,textfont=sf]{subfig}
\usepackage{textcomp}
\usepackage{verbatim}
\usepackage{url}
\newcolumntype{C}[1]{>{\centering\arraybackslash}p{#1}}
\def\BibTeX{{\rm B\kern-.05em{\sc i\kern-.025em b}\kern-.08em
    T\kern-.1667em\lower.7ex\hbox{E}\kern-.125emX}}
\usepackage{balance}

\usepackage{xcolor} % For color support

\title{\LARGE \bf
 StixelNExT++: Lightweight Monocular Scene Segmentation and\\Representation for Collective Perception
}

\author{Marcel Vosshans$^{1,2}$, Omar Ait-Aider$^{1}$, Youcef Mezouar$^{1}$ and Markus Enzweiler$^{2}$% <-this % stops a space
\thanks{$^{1}$The authors are with the Institut Pascal ISPR (Image, Systems of Perception, Robotics), Universite Clermont Auvergne INP / CNRS, France
        {\tt\small \{youcef.mezouar, omar.ait-aider\}@uca.fr}}%
        \thanks{$^{2}$The authors are with the Institute for Intelligent Systems at Esslingen University of Applied Sciences in Germany
        {\tt\small \{marcel.vosshans, markus.enzweiler\}@hs-esslingen.de}}%
}

\begin{document}
\bstctlcite{IEEEexample:BSTcontrol}
\maketitle
\thispagestyle{empty}
\pagestyle{empty}

%%%%%%%%%%%%%%%%%%%%%%%%%%%%%%%%%%%%%%%%%%%%%%%%%%%%%%%%%%%%%%%%%%%%%%%%%%%%%%%%
\begin{abstract}
%In this work, we present a novel approach to scene representation by revisiting the established concept of the Stixel World. 
%Our method filters monocular image scenes to extract the desired essence, such as obstacles in 3D space, leveraging only the intrinsics of the camera. 
%We provide resources to map a straightforward workflow: define the objects of interest, generate automatic ground truth from LiDAR data, and train our lightweight network.
%The predicted Stixels achieve high levels of information compression while maintaining adaptability to point cloud, bounding box, or BEV computations. Notably, the flexibility of using clusters of 3D Stixel units, rather than single large bounding boxes, supports use cases such as general obstacle detection and collective perception for monocular cameras.
%We compared our method against 3D bounding boxes from the Waymo dataset, achieving satisfying results for objects within \(30\) m at inference times as low as \(10\) ms per frame.

This paper presents StixelNExT++, a novel approach to scene representation for monocular perception systems.
Building on the established Stixel representation, our method infers 3D Stixels and enhances object segmentation by clustering smaller 3D Stixel units. 
The approach achieves high compression of scene information while remaining adaptable to point cloud and bird's-eye-view representations. 
Our lightweight neural network, trained on automatically generated LiDAR-based ground truth, achieves real-time performance with computation times as low as 10 ms per frame. 
Experimental results on the Waymo dataset demonstrate competitive performance within a 30-meter range, highlighting the potential of StixelNExT++ for collective perception in autonomous systems.
\end{abstract}
%%%%%%%%%%%%%%%%%%%%%%%%%%%%%%%%%%%%%%%%%%%%%%%%%%%%%%%%%%%%%%%%%%%%%%%%%%%%%%%%

%%%%%%%%%%%%%%%%%%%%%%%%%%%%%%%%%%%%%%%%%%%%%%%%%%%%%%%%%%%%%%%%%%%%%%%%%%%%%%%%
% Main content of the paper
\input{sections/1_Introduction}
\input{sections/2_RelatedWork}
\input{sections/3_StixelNExTPro}
\input{sections/4_Experiments}
\input{sections/5_Discussion}
\input{sections/6_Conclusion}

%%%%%%%%%%%%%%%%%%%%%%%%%%%%%%%%%%%%%%%%%%%%%%%%%%%%%%%%%%%%%%%%%%%%%%%%%%%%%%%%

\addtolength{\textheight}{-0.1cm}

\bibliographystyle{IEEEtran}
\bibliography{references, options}

\begin{IEEEbiography}[{\includegraphics[width=1in,height=1.25in,clip]{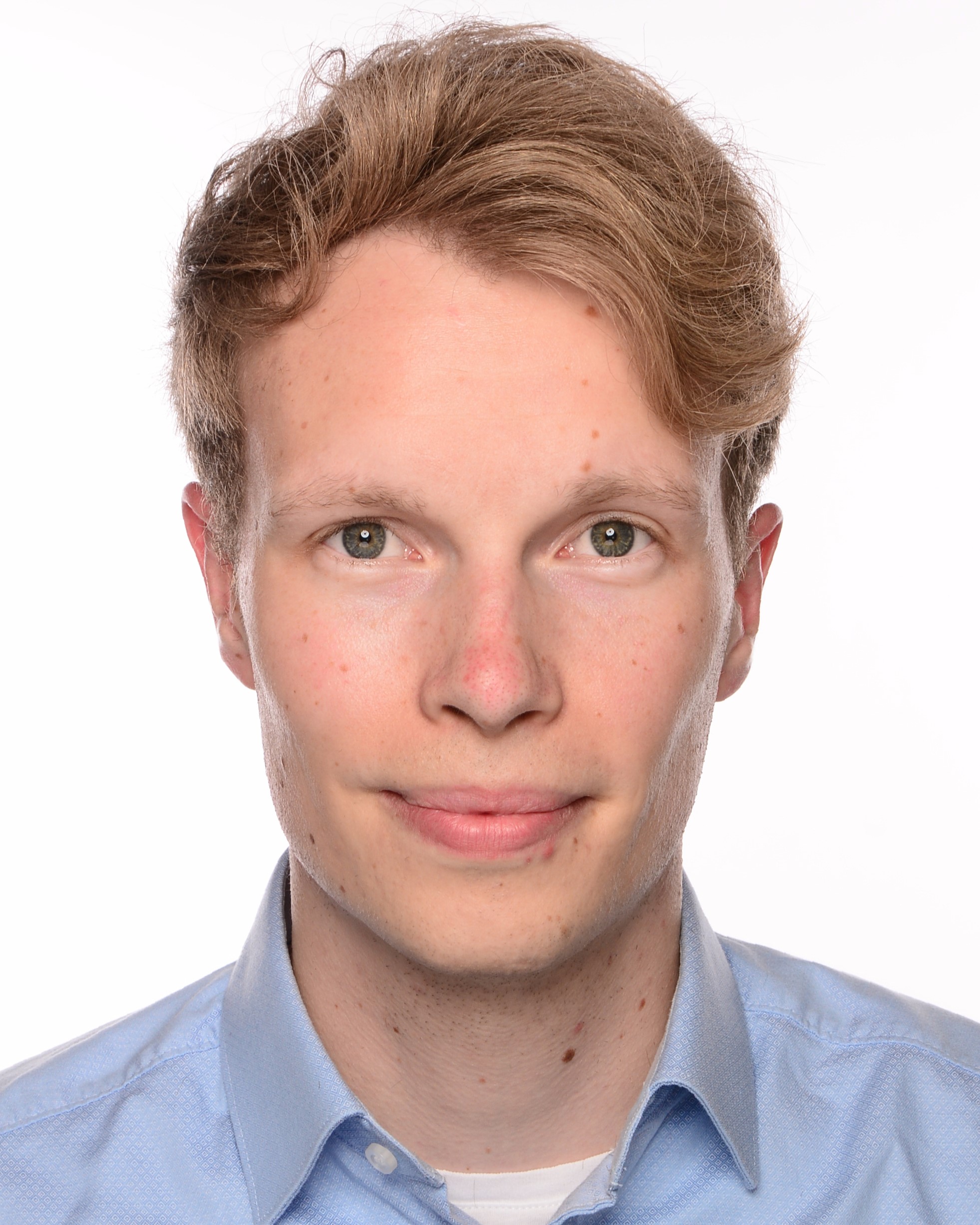}}]{Marcel Vosshans}
received his Bachelor’s degree in Electrical Engineering and Information Technology from the University of Applied Sciences Hannover in 2018. 
He completed his Master’s degree in Applied Computer Science at Esslingen University of Applied Sciences, Germany, in 2020.
Currently, he is pursuing a Ph.D. at Institut Pascal-CNRS in France, while on-site at the Institute for Intelligent Systems at Esslingen UAS. His research focuses on collective perception for autonomous public transport.
\end{IEEEbiography}
\begin{IEEEbiography}[{\includegraphics[width=1in,height=1.25in,clip]{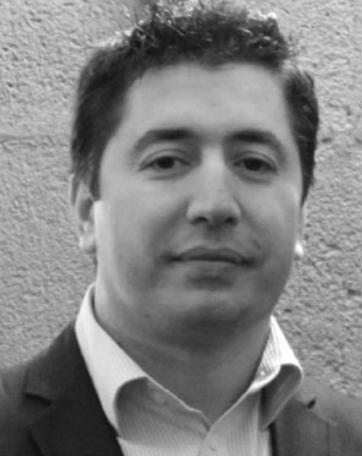}}]{Omar Ait-Aider}
is an associate professor and the head of the bachelor degree in mechatronics at University of Clermont-Auvergne. 
He received his master’s Degree on “Autonmous Robotics” at Pierre et Marie Curie University in Paris, then his PhD on “Computer Vision for Robotics” at the University of Evry Val d’Essonne in France. 
Since September 2006, he is member of the “Image, Perception Systems and Robotics” group within the Institut Pascal-CNRS. 
His research is focused on geometrical camera modelling and 3D vision.
\end{IEEEbiography}
\begin{IEEEbiography}[{\includegraphics[width=1in,height=1.25in,clip]{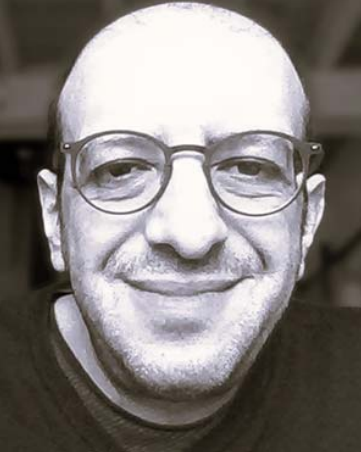}}]{Youcef Mezouar}
received the Ph.D. degrees in automation and computer science from the University of Rennes 1, France in 2001. 
He obtained the Habilitation Degree from Universite Blaise Pascal, ´Clermont-Ferrand, France, in 2009.
He spent one year as Postdoctoral Associate in the Robotic Lab of the Computer Science Department of Columbia University, New York.
He was assistant professor from 2002 to 2011 in the Physics Department of Blaise Pascal University, Clermont-Ferrand, France.
He is a Full Professor at Clermont Auvergne INP-SIGMA’Clermont since 2012.
\end{IEEEbiography}
\begin{IEEEbiography}[{\includegraphics[width=1in,height=1.25in,clip]{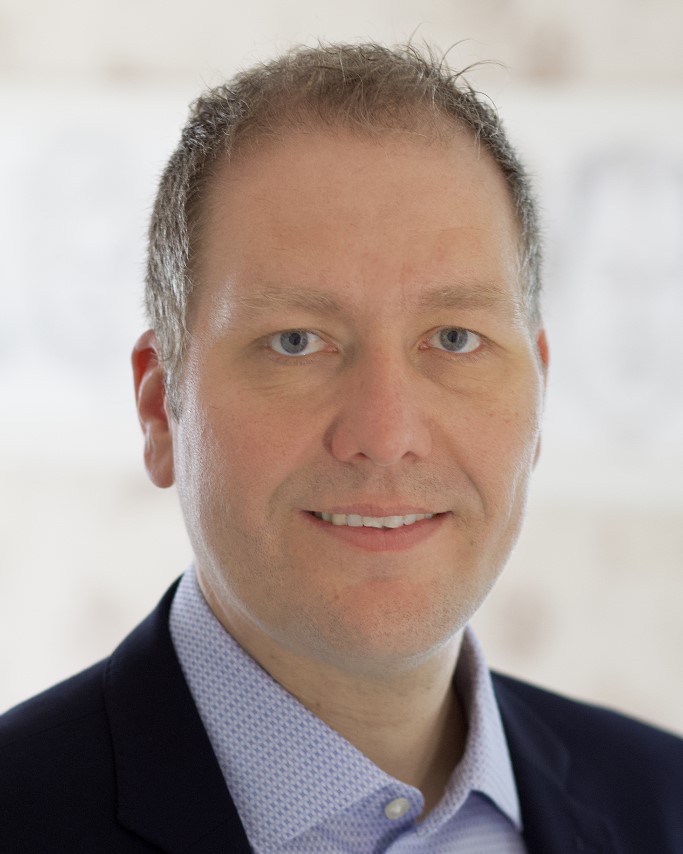}}]{Markus Enzweiler} received the Ph.D. degree in Computer Science from the University of Heidelberg in 2011.
From 2010 to 2020, he has been working with Mercedes-Benz Research \& Development in Stuttgart focusing on camera- and LiDAR-based scene understanding for self-driving cars. 
Since 2021, he is a professor of Computer Science at Esslingen University of Applied Sciences, Germany, where he founded and heads the Institute for Intelligent Systems.
\end{IEEEbiography}

\end{document}

%% file: sections/1_Introduction.tex
\section{Introduction}
Monocular imaging techniques often utilize representations such as 3D bounding boxes, occupancy grids, or pixel-wise scene completion with depth. 
While 3D bounding boxes provide a coarse, object-centered abstraction, pixel-wise scene completion enriches the representation by adding a depth channel to an RGB image, offering a finer and more generalized depiction of scene information.
However, both approaches have limitations. 
Pixel-wise depth maps, though detailed, demand significant storage space and increase bandwidth requirements during data transmission.
Conversely, bounding boxes offer a compact, preprocessed representation but lack granularity. They are effective for dynamic objects, such as vehicles, pedestrians, and bicycles, yet struggle to accurately represent larger, static elements.

In this paper, we propose a novel representation that bridges the gap between bounding box clustering and pixel-wise segmentation. 
Instead of representing large objects, such as buildings, with big bounding boxes, we subdivide these into numerous smaller boxes, clustering them to form a more detailed representation of the object. 
To achieve this, we adopt a concept that is already well-established: the Stixel \cite{denzler_stixel_2009}. Originally developed for stereo vision, Stixels are vertically oriented sticks used to represent object parts and effectively process captured scenes, forming what is known as the Stixel World.

\begin{figure}
    \centering
    \includegraphics[width=\linewidth]{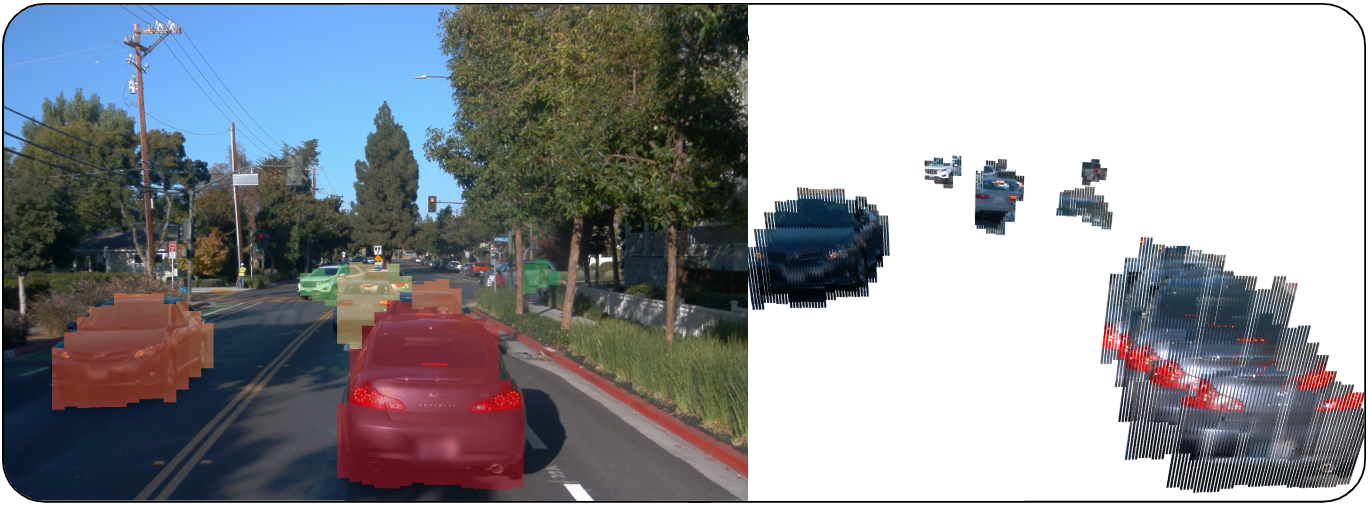}
    \caption{A \textbf{StixelNExT++ result} is depicted on the left in the 2D Stixel World Space, where relative depth is color-coded from red (near) to green (far). On the right, the corresponding prediction is visualized in 3D space with absolute metric depth, while maintaining the relative RGB values for enhanced comparison.}
    \label{fig:Stixel_World_result}
\end{figure}

Our approach enhances the traditional bounding box representation by integrating benefits commonly associated with segmentation, including improved detail and shape accuracy, the ability to represent both dynamic and static objects, and a strong emphasis on runtime efficiency and adaptability. As illustrated in Figure \ref{fig:Stixel_World_result}, the proposed StixelNExT++ neural network generates a 3D scene representation directly from monocular RGB images in an end-to-end manner.

The network is trained using ground truth data automatically generated from dense LiDAR annotations, which are adapted to include 3D bounding boxes, semantic segmentation classes, and holistic obstacle categorization. This approach enables the network to learn the spatial and geometric properties of scene elements by leveraging a depth-aware Weighted Binary Cross-Entropy (WBCE) loss for classification and regression tasks. By training on this enriched data, the network produces a flexible representation that supports generalized obstacle detection, referred to as the holistic approach.

The broader motivation for this work is to create a lightweight and adaptable representation for monocular perception systems that minimizes bandwidth requirements. Such a representation leverages existing urban camera infrastructure to facilitate collective perception for autonomous vehicles, with particular relevance for public transportation systems like buses. To further support research in this domain, we previously announced the infrastructure-synchronized AEIF-Dataset \cite{vosshans_aeif_2024} for collective perception. 

Building on our previous work introducing a LiDAR-based Stixel generator~\cite{vosshans_stixelnext_2024}, this paper presents significant advancements, including:
\begin{enumerate}
    \item \textbf{Enhanced adaptive scene representation} using Stixels to bridge the gap between bounding boxes and pixel-wise segmentation:
    \begin{itemize}
        \item Lightweight, real-time representation suitable for collective perception
        \item Improved object detail, capturing finer shapes and contours beyond traditional bounding boxes
    \end{itemize}
    \item \textbf{End-to-end neural network} for predicting a 3D Stixel World directly from monocular RGB images, featuring:
    \begin{itemize} \item Automatically generated training data and pre-trained model weights
        \item Depth-aware Weighted Binary Cross-Entropy (WBCE) loss for enhanced depth estimation
        \item Optional clustering capability for object grouping
    \end{itemize}
    \item We publicly release the \textbf{StixelNExT++ model}\footnote{\url{https://github.com/MarcelVSHNS/StixelNExT_Pro}}, along with associated datasets, the Stixel generator \cite{vosshans_stixelnext_2024}, and \textbf{pyStixel-lib}\footnote{\url{https://pypi.org/project/pyStixel-lib/}}, a DevKit for Stixel-related tasks.
\end{enumerate}

%% file: sections/2_RelatedWork.tex
\section{Related Work}
\begin{figure*}[ht]
    \centering
    \includegraphics[width=0.9\textwidth]{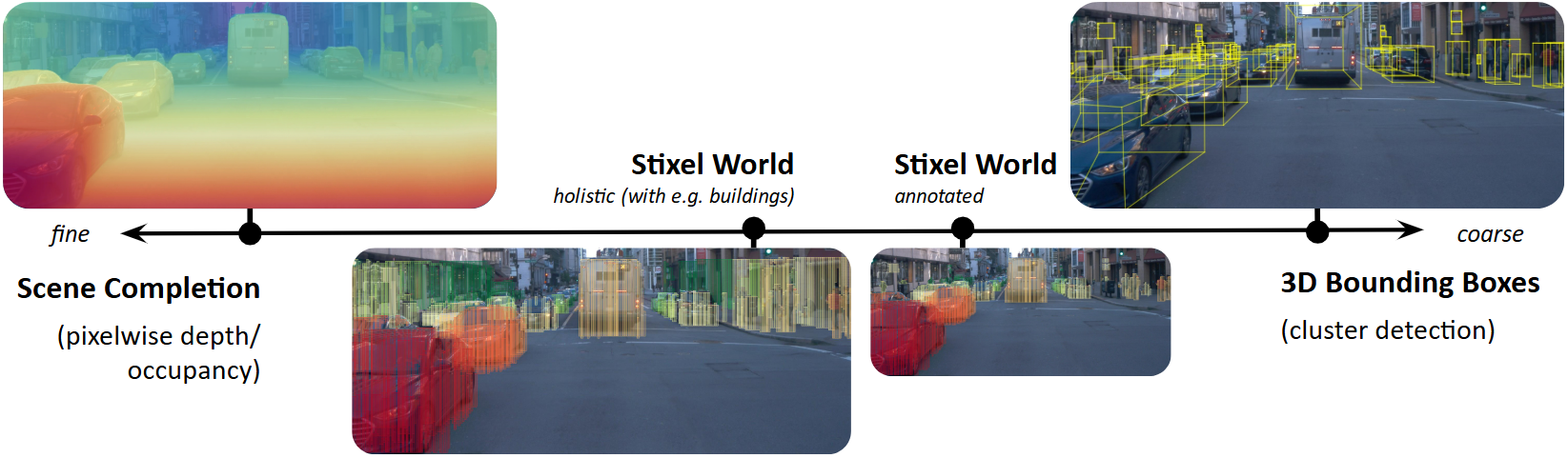}
    \caption{The \textbf{Stixel World representation spectrum} illustrates different levels of scene abstraction for robotics perception tasks. At the coarse end, 3D bounding boxes offer compact, object-level abstractions optimized for cluster-based detection. At the fine end, pixel-wise depth or occupancy maps provide detailed, high-resolution scene representations. Stixel World adaptations occupy an intermediate position, balancing detail and efficiency. Annotated Stixel Worlds are derived from 3D bounding boxes, while the holistic approach generalizes to capture all obstacles, including static structures such as buildings.}
    \label{fig:classification}
\end{figure*}
\textbf{Stixel World.}
The fundamentals of the Stixel World \cite{denzler_stixel_2009} were originally introduced as a highly effective environmental representation for stereo vision data.
Over the years, numerous advancements have been made, including multi-layer representations \cite{pfeiffer_towards_2011}, semantic \cite{schneider_semantic_2016} or e.g. instance-aware \cite{hehn_instance_2019} extensions — all achieved with stereo cameras. 
In the context of monocular 2D Stixel generation, neural networks can effectively replace the need of the second camera. 
In early works, such as StixelNet \cite{levi_stixelnet_2015}, Stixels were utilized for general obstacle detection. 
This was later extended \cite{garnett_real-time_2017} with a category-based single-shot detector and pose estimator.
In previous work \cite{vosshans_stixelnext_2024} we contributed by introducing multi-layer Stixel representations for monocular 2D obstacle detection learned from LiDAR data.
In 3D, one noteworthy approach is Mono-Stixel \cite{brickwedde_mono-stixels_2019}, which adhered more closely to the original concept of stereo-based Stixels.
It derived a Stixel World using dense optical flow fields, pixel-wise semantic segmentation, and camera motion as inputs.
In subsequent work \cite{brickwedde_exploiting_2019}, these inputs were estimated independently through neural networks, further advancing the applicability of the Mono-Stixel approach.
To train a neural network, ground truth data is often required, particularly in supervised approaches. 
In this context, Stixels can be derived from LiDAR data \cite{vosshans_stixelnext_2024,levi_stixelnet_2015} to serve as ground truth.

\textbf{Monocular Depth Estimation.}
We position our 3D Stixel World representation as an intermediate approach between a pixel-wise segmentation and bounding box clustering. 
One of the key challenges in predicting 3D Stixels end-to-end is monocular depth estimation. 
As illustrated in Figure \ref{fig:classification}, we tackle depth completion along a fine-to-coarse axis. From a fine segmentation perspective, Monodepth \cite{godard_unsupervised_2016} and its subsequent developments \cite{godard_digging_2018} serve as an early foundation, demonstrating the capability to learn depth or disparity using a CNN trained on stereo data and infer depth from a monocular image, assuming known intrinsic and extrinsic (for training) camera parameters.
While Monodepth offers real-time capability, the latest monocular depth estimators, such as Depth Anything \cite{yang_depth_2024-1}, achieve this only to a limited extent.
The adoption of foundation models in combination with transformer backbones has elevated monocular depth estimation to new levels of precision and generalization, albeit at the expense of runtime performance. 
For instance, even with the updated DepthAnything v2 \cite{yang_depth_2024}, inference times remain around 60 ms on an NVIDIA V100. 
While monocular depth estimation or scene completion operates on a pixel-wise depth map with a 1:1 ratio, monocular occupancy estimation employs a fixed 3D grid to predict whether each voxel is occupied or free. 
One example, MonoScene \cite{cao_monoscene_2021}, uses optical rules and geometric constraints within the frustum to determine depth.
Alternatively, other approaches directly learn occupancy \cite{peng_learning_2023} in 3D space, bypassing frustum depths to simplify the process and broaden applicability.

In contrast to fine pixel-wise determination, 3D bounding box detection focuses on identifying pre-clustered objects, such as vehicles, and estimating their spatial positions. 
As a baseline, we started with MonoGRNet \cite{qin_monogrnet_2020}, which used geometry constraints (similar to MonoScene) along with 2D projections to detect amodal 3D bounding boxes. 
However, across different approaches, we identified recurring challenges: the varying sizes of objects and the coverage limitations of learned object classes.
YOLOBU \cite{xiong_you_2023} addressed the issue of determining the correct center by positioning objects bottom-up, a definition we also used for our Stixel Interpretation. 
An alternative method for predicting bounding boxes involves bird's-eye-view (BEV) segmentation \cite{saha_translating_2022}, where the position of an object along the y-axis of the image significantly influences its estimated depth. For example, SeaBird \cite{kumar_seabird_2024} utilized BEV segmentations transformed from 2D image features to enhance the detection of larger objects such as buildings.

All of these approaches treat depth estimation as a regression problem.
For instance, QAF2D \cite{ji_enhancing_2024} generates bounding box anchors based on 2D bounding box detections.
Pixel-wise depth prediction leverages classification techniques \cite{cao_estimating_2018}, but the underlying challenge arises from the standard itself: it attempts to achieve fine segmentation using a coarse representation (depth classes).

\textbf{Metric Depth.}
When depth is well-predicted, it is often in the form of relative depth, representing the ratio between the depths of different areas in the image.
Using known camera intrinsics, the 3D position can be reconstructed, a process commonly referred to as pseudo-LiDAR \cite{wang_pseudo-lidar_2019}. 
Works like CAM-Convs \cite{facil_cam-convs_2019} have incorporated camera parameters to learn calibration-aware patterns, enabling more precise depth estimation. 
Another approach involves embedding the focal length directly into the model \cite{he_learning_2018}. 
They proposed a method to generate synthetic datasets with varying focal lengths derived from fixed-focal-length datasets.

%% file: sections/3_StixelNExTPro.tex
\section{STIXELNEXT++}
Scene representation forms the foundation of the perception pipeline in autonomous systems, encompassing formats like point clouds, images, or collections of bounding boxes. 
The choice of representation plays a pivotal role in determining both the precision \cite{wang_pseudo-lidar_2019} and runtime efficiency of subsequent processing stages. 
The Stixel World offers an effective trade-off between dense depth images and bounding boxes, as illustrated in Figure \ref{fig:classification}.
Its scalability arises from the flexibility to define rules for Stixel generation—for example, based on semantic criteria, such as focusing exclusively on vehicles, or geometric characteristics, such as assigning Stixels to regions with significant gradients.
When combined with a neural network, Stixels provide a mechanism to distill a scene into its most essential components, tailored to the specific needs of the application.

\textbf{Automatic Ground Truth.} 
We developed an automatic ground truth generator based on LiDAR data \cite{vosshans_stixelnext_2024}, which we specifically adapted for this purpose. 
Our updated approach now supports ground truth generation from annotations, including 3D semantic segmentation classes and 3D bounding boxes, as well as an enhanced holistic method that categorizes the environment into obstacles and free space. 
Obstacles are identified using the gradient along the z-axis in comparison to ground segmentation inliers, which include elements such as sidewalk curbs, buildings, vehicles, or vegetation. 
Figure \ref{fig:classification} illustrates a Stixel World ground truth generated using both annotation-based and holistic rules.
Compared to \cite{vosshans_stixelnext_2024}, we redefined the Stixel representation to focus exclusively on the visible parts of an object, disregarding its relation to the ground.
To achieve this, we leveraged Patchwork++ \cite{lee_patchwork_2022} for ground segmentation instead of RANSAC, using the detected outliers to derive the Stixel \cite{vosshans_stixelnext_2024}. 

\subsection{Baseline Architecture}
As previously discussed, the choice of representation plays a critical role in perception systems. The Stixel abstraction is designed to bridge the gap between raw image data and clustered objects, acting as an intermediate processing step to enhance the overall computation.
\begin{figure}
    \centering
    \includegraphics[width=\linewidth]{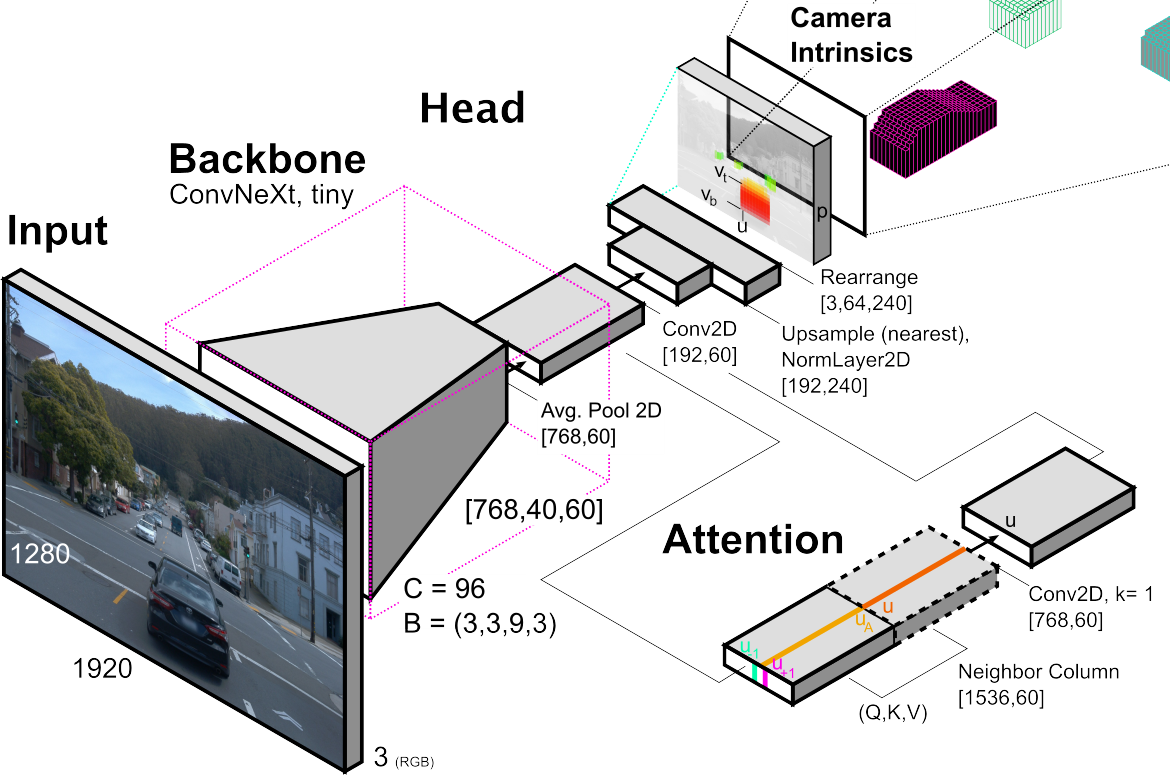}
    \caption{\textbf{The concept of StixelNExT++} involves processing a single monocular RGB image, directly at full resolution, using a pretrained ConvNeXt \cite{liu_convnet_2022} backbone. A lightweight head is attached to shape the extracted features and interpret the predictions. The network is designed to directly learn a Stixel representation, leveraging known camera intrinsics to project the Stixels into 3D space. Additionally, we investigated a column cross-attention mechanism that refines the probability of each column by incorporating information from neighboring columns.}
    \label{fig:archi}
\end{figure}

Our baseline builds upon a consistent evolution and adaptation of our 2D model \cite{vosshans_stixelnext_2024}.
This baseline will be further extended in the experimental section, where we introduce enhancements such as a depth-aware Weighted Binary Cross-Entropy (WBCE) Loss.
As illustrated in Figure \ref{fig:archi}, \textit{the input} to our system is a Full-HD monocular RGB image with dimensions \([3, 1280, 1920]\), and the ground truth data corresponds to Stixels generated using the previously described method. 
Ablations of the network architecture and runtime benchmarks will be presented in the experimental section. 
For our baseline, we utilize a pretrained ConvNeXt-Tiny \cite{liu_convnet_2022} model as the backbone. 
The architecture of the head includes an average pooling 2D layer to merge the folded height into the width, followed by dividing the depth into candidates with their respective attributes. 
To achieve the desired resolution granularity, we upscale the resulting output using nearest-neighbor interpolation, as illustrated in Figure \ref{fig:archi}. 
This design results in a vanilla network with \(27.966.624\) parameters and a passthrough size of \(\approx 6.4\) GB.

\textit{The output} of our system is a tensor with dimensions corresponding to \(240\) columns \(u\) (derived from a Stixel width of \(8\) pixels, resulting in \(64\) depth candidates, and three properties per candidate: top position \(v_{t}\), bottom position \(v_{b}\) and the probability \(P\).
This results in an output shape of \([3,64,240]\).
The network predicts the start (\(v_{t}\)) and end (\(v_{b}\)) positions of each Stixel in the image, along with the probability of being a Stixel (\(P\)). The depth candidates discretize the maximum distance into equidistant steps, ranging from \(4\) to \(66\) meters, with a step size of \(0.96\) meters. A visualization of a single column is provided below:
\includegraphics[width=\linewidth]{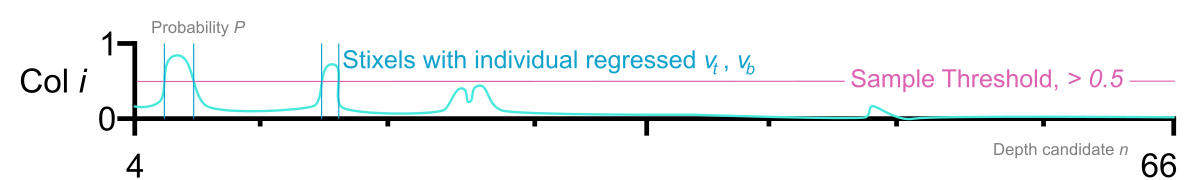}
The compromise in treating depth as a classification problem, rather than a regression problem, lies in the limitation that only one Stixel can be assigned per column and per depth candidate.
We refer to this methodology as the classification approach.

The vanilla loss function is described in Eq. (\ref{eq:total_loss}) and combines binary cross-entropy (BCE) loss \(L_{class}\) for classification and mean squared error (MSE) loss \(L_{regr}\) for regression tasks.
\begin{align}
L(p,t) = \alpha L_{regr}(v_t) + \beta L_{regr}(v_b) + \gamma L_{class}(P) \label{eq:total_loss}
\end{align}
Here, \(L\) represents the loss between the prediction \(p\) and the target \(t\), with the subscripts denoting the specific properties of the Stixel being evaluated.
This classification approach offers the advantage of simplifying the post-processing step.
For the 2D Stixel Segmentation with relative depth, techniques such as non-maximum suppression or other selection algorithms are unnecessary; we simply select Stixels that exceed the probability threshold, thereby minimizing computation time. 

In the final step of generating a 3D representation, the camera intrinsics are utilized. The fundamental equations \cite{hartley_multiple_2004} for projecting 3D points onto the image plane are given by Eqs. (\ref{eq:u}) - (\ref{eq:w}).
\begin{align}
    u &= \frac{f_x \cdot x}{z} +c_x 
    \label{eq:u} \\
    v &= \frac{f_y \cdot y}{z} +c_y 
    \label{eq:v} \\
    w &= z
    \label{eq:w}
\end{align}
The intrinsic camera parameters, represented by the matrix \(K\), include \(f_{x/y}\), the focal length in pixels, and \(c_{x/y}\), the image center in pixels. 
Since we predict \(u\), \(v\), and \(w\), the inverse camera matrix \(K^{-1}\) can be applied to compute the Cartesian coordinates, enabling the process to be efficiently executed as a matrix operation.
\begin{align}
\mathbf{P} &= \mathbf{K^{-1}} \cdot 
\begin{array}{c|c}
(\mathbf{R} & \mathbf{t})
\end{array}
\label{eq:projection} \\
\begin{bmatrix}
x \\ y \\ z \\ 1
\end{bmatrix}
&= \mathbf{P} \cdot
\begin{bmatrix}
u \cdot w \\ v \cdot w \\ w \\ 1
\end{bmatrix}
\label{eq:intrinsics}
\end{align}
In Eq. (\ref{eq:projection}), \(R\) represents the rotation matrix, and \(t\) the translation vector, which together incorporate the global coordinate system. Combined, these elements form the \((3 \times 4)\) projection matrix \(\mathbf{P}\) \cite{hartley_multiple_2004}, which accurately incorporates global positions for the projection into 3D space during post-processing. The coordinates in the image plane are denoted as \(u,v,w\), while the Cartesian space is defined by \(x,y,z\). Using Eq. (\ref{eq:intrinsics}), it is possible to calculate transformations between these two spaces, enabling seamless conversion.
In summary, this approach does not incorporate focal length awareness, as the focus is on achieving precision rather than generalization, given its application in robotics. 
Using the automatic ground truth generation method, the system can be retrained for other camera setups with different focal lengths.

\subsection{Metrics}
\begin{figure*}[tbp]
    \centering
    \includegraphics[width=0.9\textwidth]{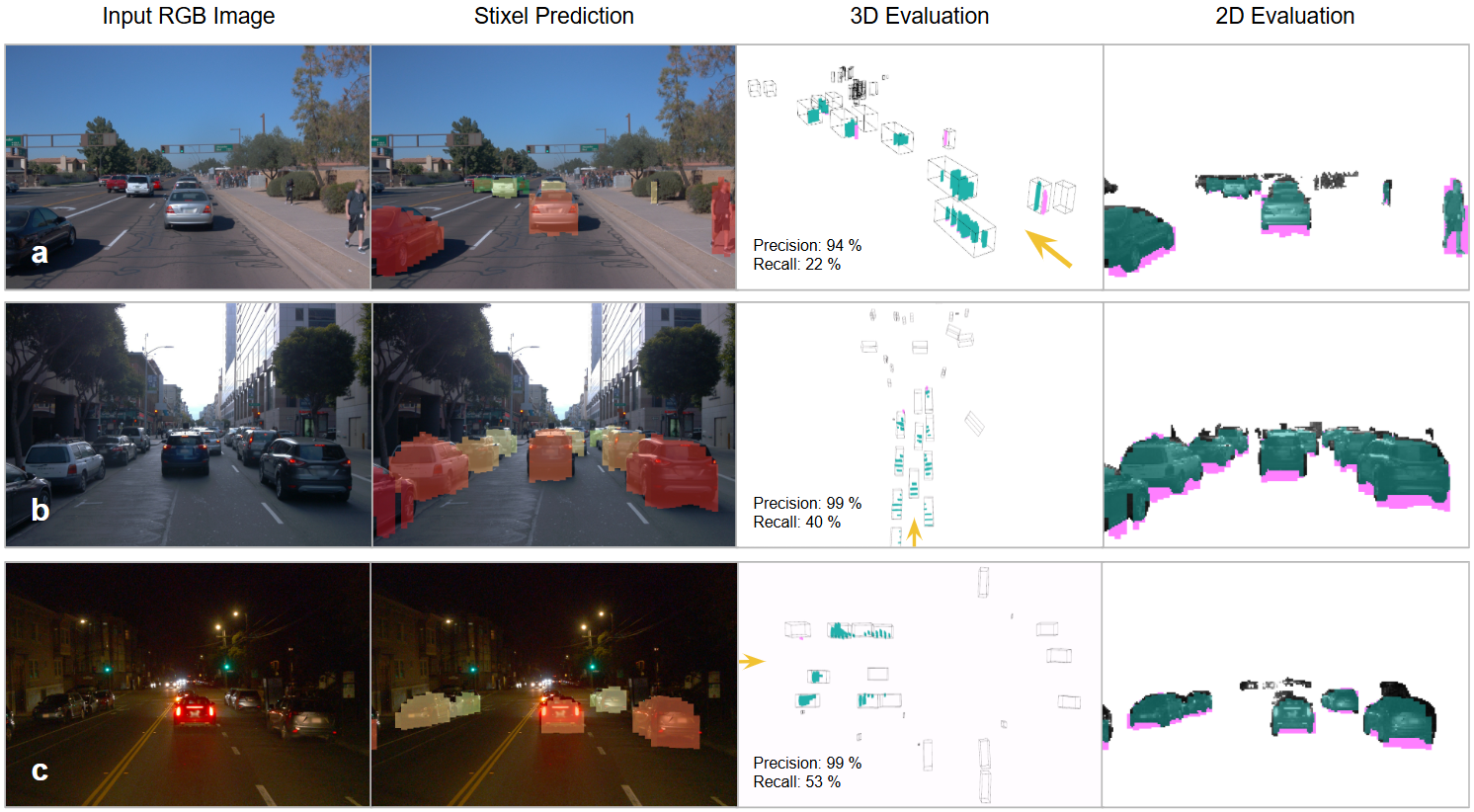}
    \caption{Sample (a-c): Qualitative inference results. Colors indicate predicted relative depth in the Stixel predictions. Yellow arrows show the field of view compared to the input image. Pink marks Stixels outside the bounding box, while turquoise marks those inside the bounding box or segmentation area. Note that many displayed boxes are not part of the evaluation. F1-Scores/Segmentation IoU are: (a) \(35\%\)/\(67\%\), (b) \(57\%\)/\(77\%\) and (c) \(69\%\)/\(66\%\).}
    \label{fig:results}
\end{figure*}
During development, we adopted a staged approach by first benchmarking our network's performance on the simpler task of 3D bounding box detection.
This allowed us to evaluate its effectiveness using established metrics while addressing well-defined geometric constraints.
The insights gained from this evaluation were then extended to the more complex, holistic Stixel-based representation.
Due to the absence of dedicated Stixel datasets or annotations, we selected a dataset that provides dense LiDAR and camera data, ultimately choosing the Waymo Open Dataset \cite{sun_scalability_2019} for its modern design, extensive size, and comprehensive LiDAR annotations.

From this dataset, we utilized the front camera, the top LiDAR, image projections, and annotations for 3D bounding boxes and 3D semantic segmentation.
For the evaluation, we combined two metrics to reflect the two steps involved in Stixel prediction.
While there are no established metrics specifically for Stixels, we aimed to align our evaluation with standardized metrics to facilitate comparisons with well-known challenges, such as 3D bounding boxes. 
Specifically, we adapted the evaluation framework from the Waymo 3D Camera-Only Detection Challenge (2022) \cite{hung_let-3d-ap_2022} and the KITTI 3D Object Detection Evaluation \cite{geiger_are_2012}.
\begin{table}[b]
\centering
\caption{Definition of the evaluated semantic classes \cite{sun_scalability_2019} for both 3D bounding box and segmentation evaluations.}
\begin{tabular}{p{0.25\linewidth}p{0.55\linewidth}}
3D Bounding Box & Panoptic Segmentation \\
\midrule
vehicle & ego-vehicle, car, truck, bus, other large vehicle, trailer \\
pedestrian  & pedestrian, pedestrian object \\
cyclist & bicycle, motorcycle, cyclist, motorcyclist \\
\midrule
\multicolumn{2}{c}{Not used} \\
\midrule
sign & sign, undefined, bird, ground animal, cone, pole, traffic light, building, road, marker, sidewalk, vegetation, sky, ground \\
\label{tab:semantics}
\end{tabular}
\end{table}
This challenge utilizes camera images and 3D bounding boxes in Cartesian space, where each bounding box is characterized by properties such as position, dimensions, heading, the number of top LiDAR points within the box, a most-visible-in-camera flag, and a semantic class (Table \ref{tab:semantics}). 
To make the results in Stixel space comparable, we introduced the following assumptions:
\begin{itemize}
    \item \textbf{Precision Calculation.} 
    A predicted Stixel, when projected into 3D space, is considered approved if the height lies within a bounding box (at least \(50 \%\)). Precision is then calculated as the ratio of approved Stixels to all predicted Stixels.
    \item \textbf{Recall Calculation.} 
    A bounding box is considered a hit if at least one Stixel lies within it. 
    While additional metrics could account for multiple Stixels within a box, this approach prioritizes safety in robotic applications, where the presence of even one Stixel indicates an obstacle to avoid.
    Recall is calculated as the ratio of hit boxes to all relevant bounding boxes.
    \item \textbf{Relevant Bounding Boxes.} 
    Since not all annotated bounding boxes are visible in the image, they are considered relevant if they:
    \begin{enumerate}
        \item Contain at least one LiDAR measurement point (indicating the object is not occluded and thus visible in the image), and
        \item Primarily appear in the front camera, following the challenge's visibility criteria.
    \end{enumerate}
    \item \textbf{Bounding Box Selection.} 
    Close objects do not always meet the criteria for relevant bounding boxes (e.g. Figure \ref{fig:Stixel_World_result}) based on the \textit{most-visible-in-front-camera} flag (as defined by Waymo annotations \cite{hung_let-3d-ap_2022}). 
    Therefore, instead of relying solely on this flag, we select all possible bounding boxes within a defined field of view (FoV) angle from the camera origin.
    Theoretically, it is possible to achieve a recall greater than \(1.0\).
    Consistent with the original challenge guidelines, we evaluate only objects within a range of up to \(75\) m \cite{hung_let-3d-ap_2022}, using the center point of the bounding box as the reference.
    \item \textbf{Empty Scene Case.} 
    If no bounding boxes are present in the image and the Stixel count is zero, both precision and recall are set to \(1.0\), as the prediction correctly identifies the absence of objects.
\end{itemize}
For reference, we implemented and trained a PGD \cite{wang_probabilistic_2021} model, a 3D bounding box detector, on the same dataset and evaluation samples using identical metrics. 
It is important to note that we consider a bounding box valid if it achieves an Intersection-over-Union (IoU) of at least \(50\%\) for each object, which is slightly more lenient than the official criteria \cite{hung_let-3d-ap_2022}.
While this comparison provides a useful reference point, it is not intended as a direct baseline, as the two approaches fundamentally differ: bounding boxes aim to enclose entire objects, whereas Stixels represent only portions of objects, emphasizing their structure and spatial distribution.
%These two approaches serve different purposes and perform distinct tasks.

The second evaluation focuses on the 2D segmentation plane and assesses the 2D regression of \(v_{t/b}\) and \(u\). 
For this evaluation, we used the Jaccard Index, commonly referred to as the PASCAL VOC IoU metric \cite{everingham_pascal_2015}. 
Leveraging the provided semantic segmentation, we filtered for classes relevant to the Waymo 3D Camera-Only Detection challenge, as outlined in Table \ref{tab:semantics}. 
Pixels corresponding to these classes are considered of interest and should be covered by the network's 2D regression. 
Conversely, pixels representing other elements (e.g., streets, buildings, trees, birds, etc.) are not part of the object shape and should not be segmented, as exemplified in Figure \ref{fig:results}.
To align with our Stixel-based approach, which defines a default Stixel width of \(8\) pixels, we sampled every 8th pixel from the semantic segmentation, resulting in a \([160, 240]\) pixel image. 
This resolution constraint also applies to the \(u\) and \(v\) predictions for each Stixel. 
For evaluation, we reconstructed an image by drawing from the top to the bottom of each predicted Stixel, yielding a similarly sized \([160, 240]\) pixel image.
By applying the PASCAL VOC IoU metric, we obtained a score that measures the 2D coverage of the objects of interest. 
The objective is to evaluate the 2D shape segmentation quality, as opposed to the 3D bounding box evaluation, which only considers whether the Stixel is inside the box without regard for the shape.
Qualitative results are illustrated in Figure \ref{fig:results}.

%% file: sections/4_Experiments.tex
\section{Experiments}
In this section, we conduct an ablation study to analyze the impact of various configurations and settings on the baseline network introduced earlier. 
Our objective is not only to improve scores but to evaluate how different design choices affect the network's performance and runtime efficiency. The classification approach, as discussed previously, aligns closely with 3D bounding boxes and contrasts with segmentation-based techniques like voxel grids (Figure \ref{fig:classification}).
As part of this analysis, we also investigate the segmentation approach, which predicts an occupancy voxel grid, similar to methods like MonoScene \cite{cao_monoscene_2021}. Voxels offer a unified structure, unlike Stixels, which vary in height. This approach involves predicting voxel pillars within a finite space and interpreting them as Stixels in a postprocessing step. While this allows for an unrestricted scene description, it produces inherently more coarse predictions when mapped back to Stixels. In contrast, the classification approach limits predictions to one Stixel per depth per column but provides a more efficient and focused representation.

Despite implementing and testing the segmentation approach, we ultimately decided to proceed exclusively with the classification approach due to its significantly lower postprocessing time.
While the segmentation approach required approximately \(1\)s per frame for postprocessing, the classification approach operates efficiently within few \(\mu\)s.
In summary, all experiments were conducted using the classification approach. 
We trained the network on \(23,691\) samples from the Waymo OD training split, utilizing all samples with LiDAR segmentations.
Evaluation was performed on \(3,151\) samples from the validation split with every second sample selected to maintain efficiency without impacting the results. 
Of these, \(385\) samples contain panoptic annotations, which were used to assess segmentation capabilities. 
For reference, we implemented PGD \cite{wang_probabilistic_2021} (a 3D bounding box detector) and trained it on the Waymo Open Dataset.
We conducted numerous additional experiments, such as adding blurring to the target matrix or varying the Stixel width.
Here, we present only the most significant results that, in our view, had the biggest impact.

\subsection{Ablation study}
We provide ablation studies on our network to validate design choices and evaluate the effectiveness of proposed methods, with the objective of further improving network performance by integrating successful approaches.
\begin{table}[tbp]
\centering
\caption{Ranking of results based on the average F1 score across all probabilities from 0.1 to 0.9 for each ablation study.}
\begin{tabular}{l c}
% \toprule
\textbf{Model} & \textbf{F1-Score} \\
\midrule
ConvNeXt, 32 Depths & $32.4\%$ \\
ConvNeXt, 64 Depths (Baseline) & $60.8\%$ \\
ConvNeXt, 128 Depths & $56.0\%$ \\
ConvNeXt, 192 Depths & $48.2\%$ \\
Swin Transformer, 64 & $59.3\%$ \\
ShuffleNetV2, 64 & $54.9\%$ \\
ConvNeXt, 64, Depth WBCE Loss & $63.4\%$ \\
\textbf{ConvNeXt, 64, WBCE, Non-Linear Discretization} & $\mathbf{63.8\%}$ \\
ConvNeXt, 64, WBCE, NLD, Attention (StixelNExT++) & $62.2\%$ \\
PGD Detector \cite{wang_probabilistic_2021} & $35.6\%$ \\
% \bottomrule
\end{tabular}
\label{tab:f1_scores}
\end{table}
\textbf{Output Shape.} \\
The number of depth anchors (or candidates) directly influences the precision of depth estimation per step. 
However, increasing this number also raises the difficulty of predicting the correct depth of a Stixel based solely on 2D feature space. 
The objective is to strike a balance between achieving sufficient depth resolution to accurately match real-world distances and avoiding excessive demands on the network, considering the information density of the available training data.
To explore this balance, we trained the baseline model with \(32\) (\(1.93\) m/step), \(64\) (\(0.96\) m/step), \(128\) (\(0.48\) m/step), and \(192\) (\(0.32\) m/step) depth candidates. 
As shown in Table \ref{tab:f1_scores}, the model with \(64\) depth anchors achieved the highest F1-Score (\(60.8 \%\)), while the model with \(32\) depth anchors showed the lowest F1-Score, with a decrease of \(-28.4 \%\). 

\textbf{Backbone.} \\
To narrow down potential backbones, we excluded networks slower than our ConvNeXt (\(29.7\) M parameters) baseline.
Based on their runtime, we evaluated several promising candidates, including EfficientNetV2 \cite{tan_efficientnetv2_2021}, MobileNetV3 \cite{howard_searching_2019}, and ShuffleNetV2 \cite{ma_shufflenet_2018}, measuring their average passthrough times (see Table \ref{tab:runtime}). 
Since our primary focus is on faster GPU-based processing, ShuffleNetV2 (\(7.5\) M) was selected for its significantly lower parameter count, alongside Swin Transformer \cite{liu_swin_2021} (\(29.5\) M), as transformer architectures are increasingly relevant despite their higher passthrough time compared to ConvNeXt.
The results are presented in Table \ref{tab:f1_scores}. 
All tested architectures performed satisfactorily, indicating a relatively small impact on overall performance across models.
We implemented all these backbones to ensure flexibility in switching architectures. 
Taking Table \ref{tab:runtime} into account, we continue with ConvNeXt due to its superior GPU performance and high evaluation scores.

\textbf{Depth Prediction.} \\
The task of a vision neural network is to extract visual features, where the Stixel position in the image plane is directly linked to these features, while depth prediction benefits only indirectly.
To enhance depth prediction, we conducted two experiments:
\subsubsection{Modified Classification Loss. \(+2.6 \%\)} 
We observed strong network performance for close objects; however, the prediction probability decreased significantly with increasing distance. Building on the BCE loss, we aimed to improve predictions for more distant objects by increasing the loss contribution for these predictions while preserving the original behavior for closer objects. To this end, we introduce a novel depth-aware Weighted Binary Cross-Entropy (WBCE) Loss and replace the \(L_{class}\) in Eq. (\ref{eq:total_loss}) with:
\begin{align}
L_{class}(p,t) &= - \frac{1}{N} \sum_{i=1}^N \alpha(d_i) \cdot BCE(p_i,t_i) 
\label{eq:wbce_loss} \\
\alpha(d_i) &= \alpha_{min} + \frac{d_i - d_{min}}{d_{max} - d_{min}} \cdot (\alpha_{max} - \alpha_{min})
\label{eq:depth_loss}
\end{align}
In Eq. (\ref{eq:wbce_loss}), we implement a linear growth factor according to the original BCE formulation over \(N\) batches, incorporating a depth-dependent multiplier (Eq. (\ref{eq:depth_loss})) through matrix multiplication. Here, \(d\) represents the depth range (\(d_i\), \(4\) - \(66\) m) and \(\alpha\) is a scaling factor for the loss (ranging from \(1x\) to \(2x\)). The output demonstrated the desired effect, achieving more consistent probabilities for distant objects. This adjustment resulted in a \(+2.6 \%\) improvement in overall performance.
\subsubsection{Depth Discretization.} 
\begin{figure}[tbp]
    \centering
    \includegraphics[width=0.96\linewidth]{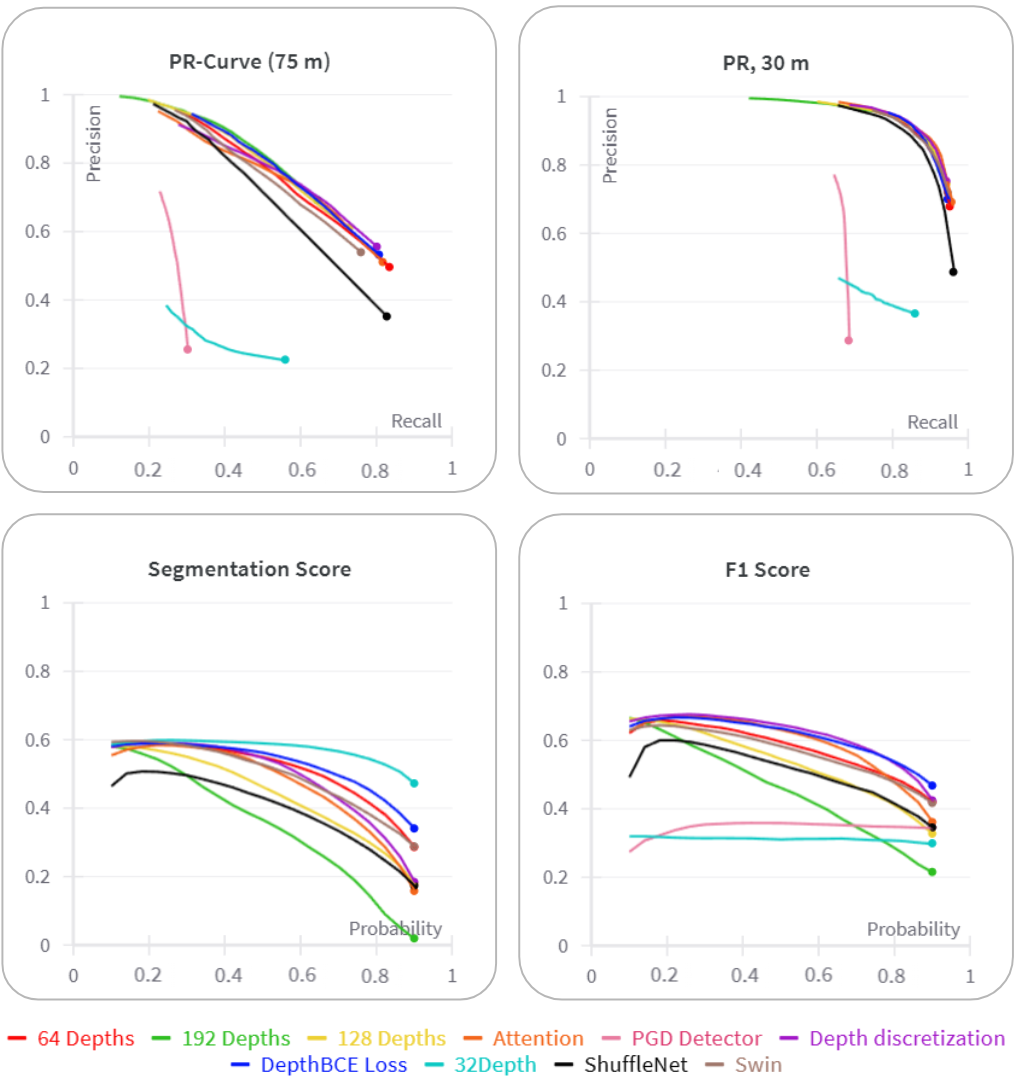}
    \caption{\textbf{PR-Curves} are presented for ranges up to 30 m and the full distance of 75 m, along with the \textbf{segmentation score} as a function of the probability threshold for classifying a Stixel. The average F1-scores are summarized in Table \ref{tab:f1_scores}. Apart from a few outliers (32 Depth and PGD), the performance differences are comparable. }
    \label{fig:quant_results}
\end{figure}

To enhance depth representation, we introduced non-linear depth anchors, providing finer depth classes for near objects and larger steps for far objects.
The discretization of depth classes can generally be described using Eq. (\ref{eq:discretization}), where \(n\) denotes the total number of classes, \(c\) represents the specific class (ranging from \(c_0\) to \(c_{max}\)), and \(a\) serves as a limiting factor for the tangent function. 
\begin{align}
D(c_i) = c_0 + \frac{\tan\left(\frac{i}{n - 1} \cdot \frac{\pi}{a}\right)}{\tan\left(\frac{\pi}{a}\right)} \cdot (c_{max} - c_0)
\label{eq:discretization}
\end{align}
Instead of using an equidistant spacing for depth bins, where the intervals between successive values remain constant (\(c_i\)), we adopted a tangential approach (\(tan(c_i)\)). This adjustment results in initial depth class differences of \(0.7\) m, increasing to differences of up to two meters at farther distances. By \(30\) m, the non-linear configuration includes \(43\) depth anchors, meaning approximately two-thirds of the depth anchors are allocated within half the maximum distance. As illustrated in Figure \ref{fig:quant_results}, the use of non-linear depth anchors resulted in a performance improvement of \(+0.4 \%\), giving the network greater flexibility in closer ranges. However, as the distance increases, prediction accuracy diminishes due to degrading probabilities, a common challenge in depth estimation.

\textbf{Attention Head.} \\
When representing an object using clustered Stixels, two vertical edges are defined from the segmentation perspective: the left and right boundaries of the object (e.g. the black car in Figure \ref{fig:archi}). 
However, many Stixels are contiguous, meaning their depth and 2D positional predictions exhibit minimal differences.
Inspired by the concept of learning occupancy \cite{peng_learning_2023}, we investigated a column cross-attention module. 
This module adds an additional layer before the network head to incorporate information from adjacent columns (left and right). 
While this design introduces a natural offset due to object edges and shapes, it aligns well with most objects, such as vehicles, buildings, cones, and signs. The attention mechanism is illustrated in Figure \ref{fig:archi}. The aggregation function combines features from the backbone and neighboring columns using concatenation, followed by a convolutional layer for integration.

In the column cross-attention module, the attention operation is inspired by the standard formulation of \(Attention(Q, K, V)\) \cite{vaswani_attention_2017}. Here, the query \(Q\) corresponds to the Stixel positions \(v_t\) and \(v_b\) and the associated depth class probabilities \(P\). The keys \(K\) represent properties of the neighboring columns \(u\), while \(V\) denotes the corresponding feature values.

Despite its conceptual advantages, the mechanism did not yield an improvement in the \textit{average} F1-Score. As shown in Figure \ref{fig:quant_results}, performance decreased for probabilities above \(70\%\). However, it performed best at our preferred operating point (up to \(30\) meters with a threshold of \(38\%\)), achieving a recall of \(89.3\%\) (compared to \(89.2\%\) for the second-best model) and a precision of \(88.3\%\) (compared to \(87.1\%\)).

\subsection{Runtime}
\begin{table}[tbp]
\centering
\caption{Inference runtimes indicating both CPU and GPU passthrough times measured on the same hardware.}
\begin{tabular}{lccc}
%\toprule
\textbf{MODEL} & \(\mathbf{t_{CPU}}\) & \(\mathbf{t_{GPU}}\) & \textbf{Parameters} \\
\midrule
ConvNeXt, tiny \cite{liu_convnet_2022}   & \(713.3\) ms & \(1.44\) ms & \(29.74\) M\\
EfficientNetV2, small \cite{tan_efficientnetv2_2021}  & \(713.2\) ms & \(4.77\) ms & \(22,2\) M\\
MobileNetV3, large \cite{howard_searching_2019} & \(219.7\) ms & \(1.75\) ms & \(4.93\) M\\
Swin Transformer, tiny \cite{liu_swin_2021}   & \(1082.4\) ms & \(4.28\) ms & \(29.5\) M\\
ShuffleNetV2, 2.0x \cite{ma_shufflenet_2018} & \(184.5\) ms & \(1.84\) ms & \(7.51\) M\\
PGD (3D BBox) \cite{wang_probabilistic_2021} & \(-\) & \(17.27\) ms & \(55.98\) M\\
%\bottomrule
\end{tabular}
\label{tab:runtime}
\end{table}
To evaluate runtime performance, we tested our various model variants on a PC running PyTorch on Ubuntu 24.04 with an NVIDIA GeForce RTX 4090 GPU, an AMD Ryzen 9 9950X CPU, and \(32\) GB of DDR5 memory at \(6000\) MHz.

For runtime measurement, we generated 1,000 random tensors with dimensions \([3,1280,1920]\), processed them one by one through the GPU for inference, and fetched the results back to the CPU for a postprocessing step, as described in the previous chapter. 
The total time was divided by the 1,000 samples to calculate the average runtime per sample. 
We compared the complete workload computation times with GPU support and without GPU support (CPU). 
The results are summarized in Table \ref{tab:runtime}.

Direct runtime comparisons between conceptually different approaches are challenging due to differences in output types — such as our Stixel image filtering compared to a 3D bounding box detector or a scene completion network. 
Nonetheless, we implemented the PGD \cite{wang_probabilistic_2021} model on our hardware to facilitate a direct comparison.

\subsection{Qualitative Results}
\textbf{Clustering.}
When discussing the Stixel World in combination with StixelNExT++, we refer to it as \textit{hybrid representation}.
This is because, on one hand, it includes computed elements resembling bounding boxes, while on the other, it retains characteristics akin to an image, lacking a definitive analysis to classify regions as objects or non-objects. 

\begin{figure}[!t]
    \centering
    \includegraphics[width=\linewidth]{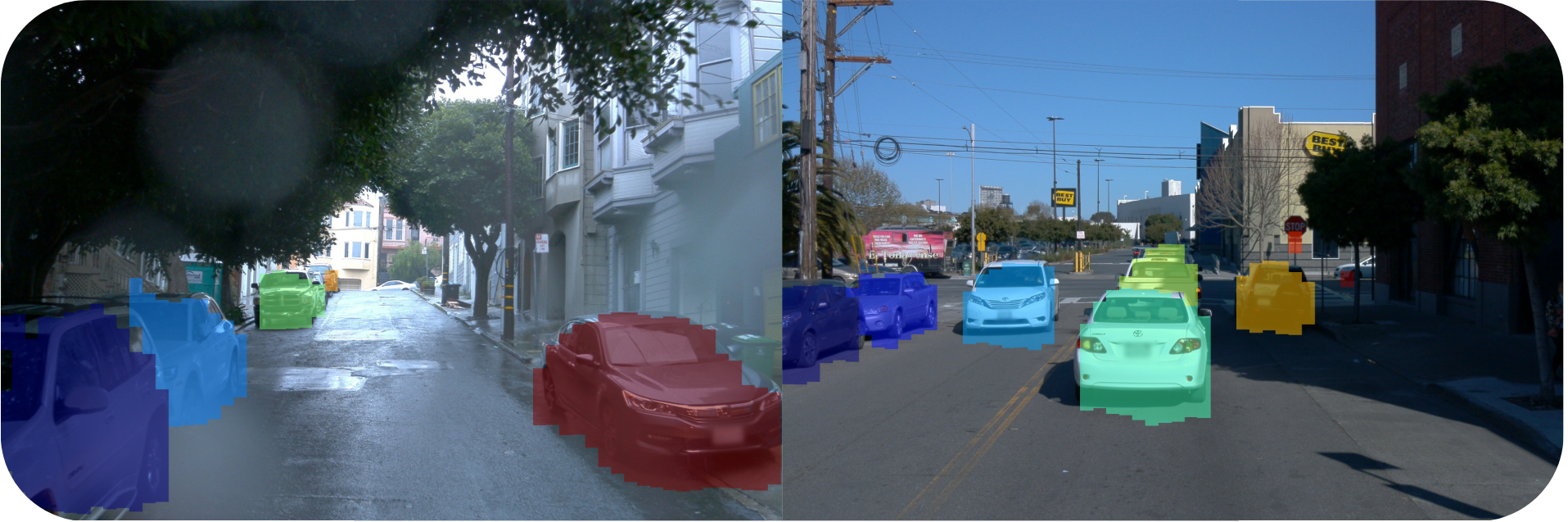}
    \caption{Two \textbf{clustering results} of a predicted Stixel World using the DBSCAN algorithm. Both examples showed promising outcomes with CPU runtimes of \(16.5\) ms (left) and \(17.1\) ms (right), respectively.}
    \label{fig:clustering_result}
\end{figure}

One notable strength of Stixels lies in their ability to augment an existing Stixel World with information from another traffic agents.
However, to obtain a representation that is somewhat comparable to a 3D bounding box detector, it is necessary to cluster the Stixels into objects. While this work does not focus on developing a dedicated Stixel clustering algorithm, we provide illustrative results using DBSCAN clustering in Figure~\ref{fig:clustering_result}. Although object dimensions can be derived from Stixel clusters, directly comparing these results to bounding boxes remains ill-posed, as the two approaches serve fundamentally different purposes.

\textbf{Holistic training.}
The primary objective was to develop a data-driven model without relying on specifically tailored datasets. To achieve this, we validated performance using comparable tasks and transferred effective improvements from the classification approach to the holistic approach.

A key distinction between these approaches lies in ground truth generation. Patterns such as significant depth differences along the \(y\)-axis or steep gradients along the \(z\)-axis in LiDAR data were used to enable the network to learn general obstacle detection rather than object-specific categorization. Annotated bounding boxes, when used to generate ground truth, provide the advantage of assigning an individual ground plane to each object.

When deriving Stixels from LiDAR data, the process identifies features like depth gaps or \(z\)-axis gradients and extends these points downward to align with the ground plane. This method is particularly effective in scenes with annotated object ground planes, where consistent starting and ending levels across all Stixels implicitly support ground plane learning.

In a holistic approach where annotations—and therefore explicit ground planes—are unavailable, detecting a single constant ground plane per scene, as previously done, becomes impractical. While this simplifies processing, it introduces inaccuracies in hilly or uneven terrains. 

To address this, we employed Patchwork++ \cite{lee_patchwork_2022}, which segments a point cloud into ground and non-ground points, enabling a more comprehensive scene representation that includes buildings, post boxes, and unknown objects. Without a reference plane, we focused on visual feature extraction, segmenting only the visible parts of objects. For instance, a Stixel corresponding to a partially occluded building ends where the occluding object (e.g., a car) begins, converting visual artifacts into Stixels.

\begin{figure*}
    \centering
    \includegraphics[width=0.9\linewidth]{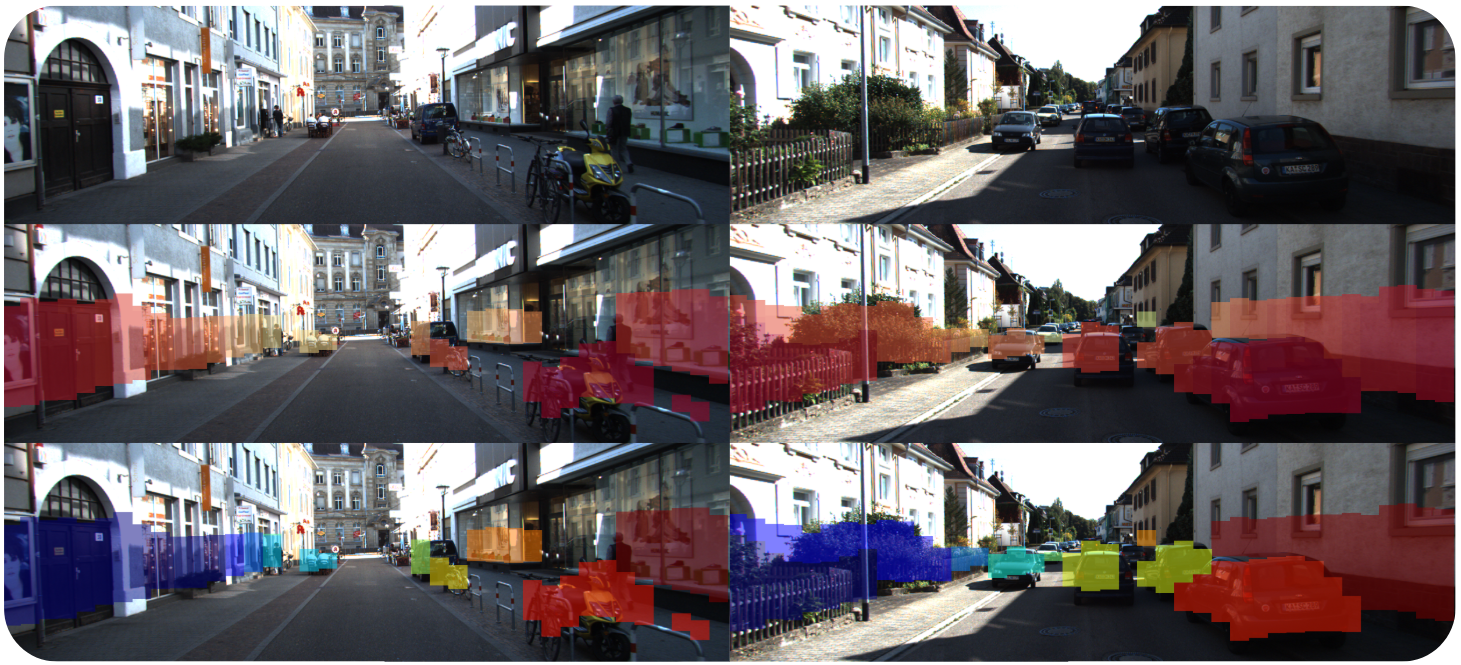}
    \caption{A qualitative result from \textbf{holistic training} on the KITTI dataset \cite{geiger_are_2012} is presented. Each sample illustrates, from top to bottom: the input image, the predicted depth Stixels, and the clustered Stixels.}
    \label{fig:holistic_result}
\end{figure*}

Due to the lack of ground truth in this setup, qualitative results using the KITTI dataset are presented in Figure \ref{fig:holistic_result}. As previously mentioned, we utilized annotated Waymo data to identify a well-performing model, which was then transferred to the holistic approach for training and deployment in a non-annotation pipeline.

%% file: sections/5_Discussion.tex
\section{Discussion and Future Work}
Evaluating a representation like Stixels poses significant challenges due to the lack of benchmarks or ground truth data for direct comparison.
As a result, we adapted existing metrics, such as those used for 3D bounding boxes or voxel grids, to provide a reference for evaluation.
However, this constitutes a slightly ill-posed comparison.

Our primary objective was to demonstrate that a learned Stixel representation can achieve comparable performance to existing technologies, such as 3D bounding boxes, while offering greater flexibility and information density.
Among various aspects of evaluation, such as semantic segmentation, we identified one key area of interest: the dependency of Stixels on camera intrinsics.
Projecting Stixels into 3D space requires knowledge of the camera intrinsics, particularly the focal length. 
The weights of StixelNExT++ are therefore optimized for a specific focal length—namely, the one used in the training dataset (Waymo). 
This dependency on focal length is a common limitation of all monocular depth estimators. 

Hence, future work involves several strategies to address this limitation. One approach is to incorporate the focal length into the prediction and ground truth data, normalizing the depth to a fixed reference  \cite{facil_cam-convs_2019}. 
Another approach is to train the network on datasets with varying focal lengths and predict the focal length alongside the depth \cite{he_learning_2018}.
Additionally, exploring focal-length-independent representations using learned transformations that align features across different camera setups could provide a promising direction for increasing model robustness.

Beyond addressing focal length dependency, future efforts will focus on developing a spatial registration strategy for collective perception based on StixelNExT++. 
To support this, we plan to utilize the AEIF dataset \cite{vosshans_aeif_2024}, which contains synchronized recordings of various scenes captured from both the ego-vehicle and infrastructure-based perception components.

%% file: sections/6_Conclusion.tex
\section{Conclusion}
In this work, we presented StixelNExT++, an end-to-end learned approach for monocular scene representation tailored to collective perception.
Building on the Stixel World concept, our method adapted the representation for monocular inputs, extending its application to general obstacle detection and scene representation. 

Our approach achieved promising performance, leveraging only a single monocular RGB image without temporal context, while maintaining real-time capabilities, with the majority of computational load attributable to a replaceable backbone network. This ensures adaptability to diverse architectures without sacrificing efficiency. By adapting the Stixel World representation, we addressed limitations of voxel- and bounding box-based representation, achieving a balance between information density and flexibility.

In doing so, StixelNExT++ establishes a robust foundation for lightweight, efficient perception in autonomous systems, offering a flexible and scalable solution for monocular scene understanding.